\pdfoutput=1

\documentclass[11pt]{article}

\usepackage{ACL2023}

\usepackage{times}
\usepackage{latexsym}
\usepackage{multirow}
\usepackage{booktabs}
\usepackage{pifont}
\usepackage{microtype}
\usepackage{latexsym}
\usepackage{booktabs}
\usepackage{courier}
\usepackage{amsmath}
\usepackage{enumitem}
\usepackage{booktabs} 
\usepackage{siunitx} 
\usepackage{subfig}
\usepackage{placeins}
\usepackage{tikz}
\usepackage{hyperref}

\usepackage[T1]{fontenc}

\usepackage[utf8]{inputenc}

\usepackage{microtype}

\usepackage{xspace}

\newcommand{\ie}{i.e.,\xspace}
\newcommand{\eg}{e.g.,\xspace}

\newcommand{\cmark}{\ding{51}}%
\newcommand{\xmark}{\ding{55}}%

\usepackage{inconsolata}
\usepackage{tcolorbox}
\definecolor{lightblue}{rgb}{0.12, 0.44, 0.80}
\definecolor{cyan}{HTML}{4BA1BF}
\definecolor{mustard}{HTML}{FEDC56}
\definecolor{sepia}{HTML}{E3B778}

\title{Edinburgh Clinical NLP at MEDIQA-CORR 2024:\\Guiding Large Language Models with Hints}

\author{
Aryo Pradipta Gema\textsuperscript{1}\thanks{\hspace{0.5em}Equal contribution.} \quad Chaeeun Lee\textsuperscript{1}\footnotemark[1] \quad Pasquale Minervini\textsuperscript{1} \quad Luke Daines\textsuperscript{2} \\ \textbf{T. Ian Simpson\textsuperscript{1}} \quad \textbf{Beatrice Alex\textsuperscript{3,4}} \\
\textsuperscript{1}School of Informatics, University of Edinburgh \quad 
\textsuperscript{2}Usher Institute, University of Edinburgh\\
\textsuperscript{3}Edinburgh Futures Institute, University of Edinburgh\\
\textsuperscript{4}School of Literatures, Languages and Cultures, University of Edinburgh\\
\texttt{ \{aryo.gema, chaeeun.lee, p.minervini, luke.daines\}@ed.ac.uk}\\
\texttt{ \{ian.simpson, b.alex\}@ed.ac.uk}
}

\begin{document}
\maketitle
\begin{abstract}

The MEDIQA-CORR 2024 shared task aims to assess the ability of Large Language Models (LLMs) to identify and correct medical errors in clinical notes.
In this study, we evaluate the capability of general LLMs, specifically GPT-3.5 and GPT-4, to identify and correct medical errors with multiple prompting strategies.
Recognising the limitation of LLMs in generating accurate corrections only via prompting strategies, we propose incorporating error-span predictions from a smaller, fine-tuned model in two ways: 1) by presenting it as a hint in the prompt and 2) by framing it as multiple-choice questions from which the LLM can choose the best correction.
We found that our proposed prompting strategies significantly improve the LLM's ability to generate corrections.
Our best-performing solution with 8-shot + CoT + hints ranked sixth in the shared task leaderboard.
Additionally, our comprehensive analyses show the impact of the location of the error sentence, the prompted role, and the position of the multiple-choice option on the accuracy of the LLM.
This prompts further questions about the readiness of LLM to be implemented in real-world clinical settings.\footnote{Our code is available at \url{https://github.com/aryopg/mediqa}}

\end{abstract}

\section{Introduction}

Medical errors represent a major concern in the healthcare sector, leading to adverse patient outcomes and higher costs for healthcare providers.
The detection and correction of such medical errors are critical in enhancing healthcare delivery and outcomes.
Recognising the importance of efficient and precise medical documentation, the MEDIQA-CORR 2024 shared task~\cite{mediqa-corr-task} is initiated to evaluate the potential of using Large Language Models (LLMs) as solutions to locate and correct medical errors within clinical notes.

In our study, we evaluated multiple prompting strategies such as In-context Learning (ICL) and Chain-of-Thought (CoT) to enhance the performance of LLMs, specifically focusing on GPT-3.5 and GPT-4~\cite{openai2023gpt4}.
We proposed incorporating a smaller fine-tuned language model, namely BioLinkBERT~\cite{yasunaga-etal-2022-linkbert}, to aid LLMs in locating an error span in a clinical note.
We incorporated the predicted error span in two ways: 1) by presenting it as a hint in the prompt to direct the error correction, and 2) by framing it as multiple-choice questions where the LLM can select the most probable correction.

Our findings revealed that the LLMs show noticeable improvements in their generation capability when presented with more ICL examples.
Similarly, the CoT prompt also improves the error correction capability of the LLMs.
Among the different reasoning styles we experimented with, the LLM performs the best with brief reasoning.
Our prompt design, which provides a hint about the typical nature of the errors and a hint from the error span prediction, further improves the LLMs' ability to generate corrections.
The combination of 8-shot ICL with Brief CoT reasoning and hints is the best-performing prompting strategy in the two provided validation sets.
This pipeline ranked sixth in the shared task leaderboard.
In summary, our contributions are as follows:
\begin{itemize}[leftmargin=*]
    \item A comprehensive analysis of the impact of ICL on the performance of LLMs for medical error correction.
    \item An extensive exploration of CoT to inject various reasoning styles into the LLM and their impact on the performance.
    \item Novel approaches to integrate the predictions of a smaller language model into the LLM generation.
    \item Sensitivity analyses of LLMs, highlighting how minor variations such as the error sentence location, the prompted role, and the multiple-choice positioning can influence generation capabilities.
\end{itemize}

\begin{figure*}[ht]
    \centering
    \includegraphics[width=\textwidth]{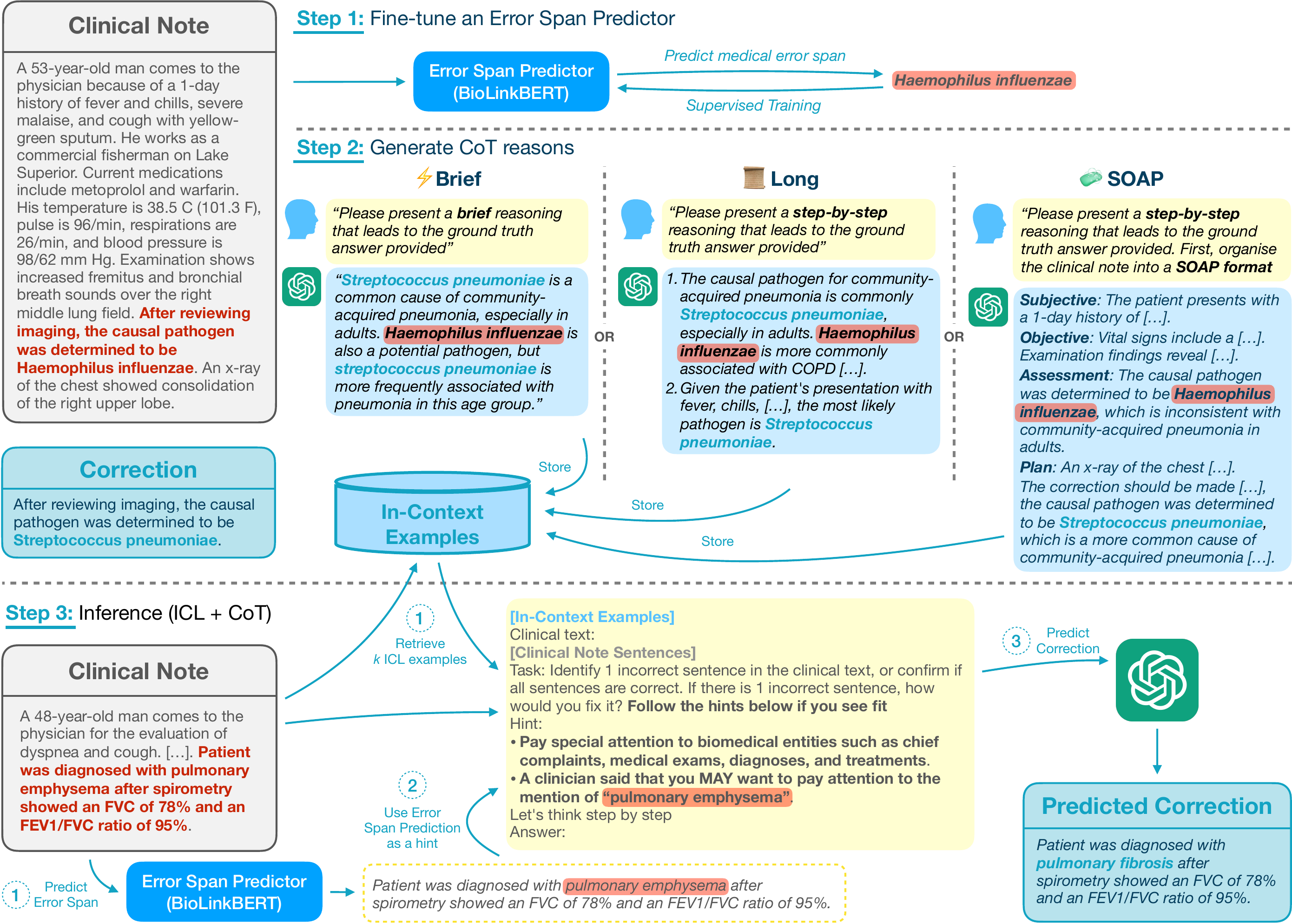}
    \caption{Schema of our best-performing strategy with In-Context Learning (ICL) and Chain-of-Thought (CoT) prompting strategies. The strategy involves fine-tuning BioLinkBERT on the training set for error span prediction. Then, we prompt GPT-3.5 with various reasoning templates to reason pairs of clinical notes and ground truth corrections to gather ICL examples with CoT reasons. Subsequently, this strategy leverages the ICL examples and error span predictions as a hint.}
    \label{fig:e2e_schema}
\end{figure*}

\section{Background}

\subsection{Task Description}

MEDIQA-CORR 2024 task~\cite{mediqa-corr-dataset} comprises three sub-tasks, each addressing a different aspect of medical error correction:
\begin{description}[style=unboxed,leftmargin=0cm,noitemsep,topsep=0pt]
    \item[Binary classification:] Detecting whether the clinical note contains a medical error.
    \item[Span Identification:] Identifying the text span associated with a medical error if it exists.
    \item[Natural Language Generation:] Generating a correction if a medical error exists.
\end{description}
Table \ref{table:data_stats} shows the statistics for each data split, organised by the source of the data and whether or not it contains a medical error. Each clinical note contains either one or no medical error.
\begin{table}
\centering
\resizebox{\columnwidth}{!}{
\begin{tabular}{lcccccc}
\toprule
              \multirow{2}{*}{Category} & \multicolumn{2}{c}{Train} & \multicolumn{2}{c}{Valid}   & \multicolumn{2}{c}{Test}    \\
\cmidrule(lr){2-3}
\cmidrule(lr){4-5}
\cmidrule(lr){6-7}
               & MS       & UW             & MS           & UW           & MS           & UW           \\
\midrule
No Error       & 970      & 0              & 255          & 80           & -            & -            \\
Contain Error  & 1,219     & 0              & 319          & 80           & -            & -            \\
\midrule
\textbf{Total} & \textbf{2,189} & \textbf{0}     & \textbf{574} & \textbf{160} & \textbf{597} & \textbf{328} \\
\bottomrule
\end{tabular}
}
\caption{Dataset statistics of each split, categorised by the source and presence of a medical error.}
\label{table:data_stats}
\end{table}

The task uses accuracy for binary classification and span identification.
The generated correction is evaluated using an aggregate Natural Language Generation (NLG) score, combining ROUGE-1~\cite{lin2004rouge}, BERTScore~\cite{zhang2020bertscore}, and BLEURT~\cite{sellam2020bleurt}, which is best aligned with human judgement, among other NLG metrics~\cite{ben-abacha-etal-2023-investigation}.

\subsection{Related work}

LLMs have shown remarkable capabilities in many NLP tasks, including in the clinical domain.
\citet{lievin2022can} evaluated LLMs with various prompting strategies, showing LLMs' capability to answer complex medical questions.
\citet{falis2024can} uses GPT-3.5 to generate accurate synthetic discharge summaries by prompting it with a list of diagnoses.
\citet{gema2024edinburgh} also shows GPT-4 in zero-shot setting outperforms other fine-tuned LLMs in a natural language inference task for clinical trial data.

However, despite the increasing use of general LLMs, their performance varies widely depending on the nature of the task.
For instance, fine-tuned smaller encoder-based models (\eg BioLinkBERT) still maintain the lead in tasks such as medical entity recognition~\cite{kim2023exploring}.
\citet{gema2023parameter} showed that domain-adapted LLaMA~\cite{touvron2023allama} outperforms the state-of-the-art models in clinical outcome prediction tasks.
Such studies show that fine-tuned models are still preferable, especially in discriminative tasks such as classification and entity recognition.

In this study, we seek to combine the generative capability of LLMs with the discriminative capability of a smaller fine-tuned language model.
We compared our novel method with solutions that rely solely on prompting strategies (\ie ICL and CoT).

\section{System Overview}

We experimented with three strategies:
%

\begin{description}[style=unboxed,leftmargin=0cm,noitemsep,topsep=0pt]
    \item[End-to-end Prompting Strategy for Error Correction:] This strategy treats all three subtasks as a single prompting task. The LLM simultaneously predicts if the clinical note contains an error, pinpointing its location, and proposing a correction.
    \item[Fine-tuning Error Span Prediction and MCQ- style Error Correction:] This method splits the task into error span prediction and correction. It uses a fine-tuned model for error span prediction, followed by MCQ-style prompts for correction.
    \item[Hybrid Approach:] As shown in Figure~\ref{fig:e2e_schema}, This approach uses error span predictions from a fine-tuned model as correction hints injected into the end-to-end prompting strategy. This is our best-performing strategy in both validation and test sets.
\end{description}

%


The following sections outline the details for the \textbf{Error Span Prediction} and \textbf{Error Correction}.

\subsection{Error Span Prediction}\label{sec:err_span_pred}

We noticed that medical errors appear predominantly in the form of diagnoses or treatments, instead of the patient’s factual information.
This finding motivated us to fine-tune an encoder model to first detect an error span within the clinical note.

We trained BioLinkBERT and BERT\footnote{Both base and large versions of the models} using a question-answering pipeline adapted from the Stanford Question Answering Dataset (SQuAD).
We pre-processed the training and validation sets to align them with the SQuAD v1 format, which assumes that there is always an error span in the input.
We introduced a template question, ``Which part in the given clinical note is clinically incorrect?'' in the question column of the SQuAD format.
The trained model predicts the start and end indices, which indicate the position of the predicted error span in the text. 

We trained and evaluated the error span prediction models only on clinical notes that contained errors.
We evaluated the models using exact match (EM) and token-based F1 score metrics, using the latter to choose the best checkpoint.



\subsection{Error Correction}

We experimented with GPT-3.5 and GPT-4 for the error correction step.
We prompted the LLMs to return the outputs in JSON format for ease of postprocessing.
In rare cases where the outputs are not JSON-parseable, we default the prediction as if no error was found.
We integrated the error span prediction to this error correction step in two ways:

\subsubsection{Multiple-Choice Question prompt}

\begin{figure}[t]
    \centering
    \includegraphics[width=\columnwidth]{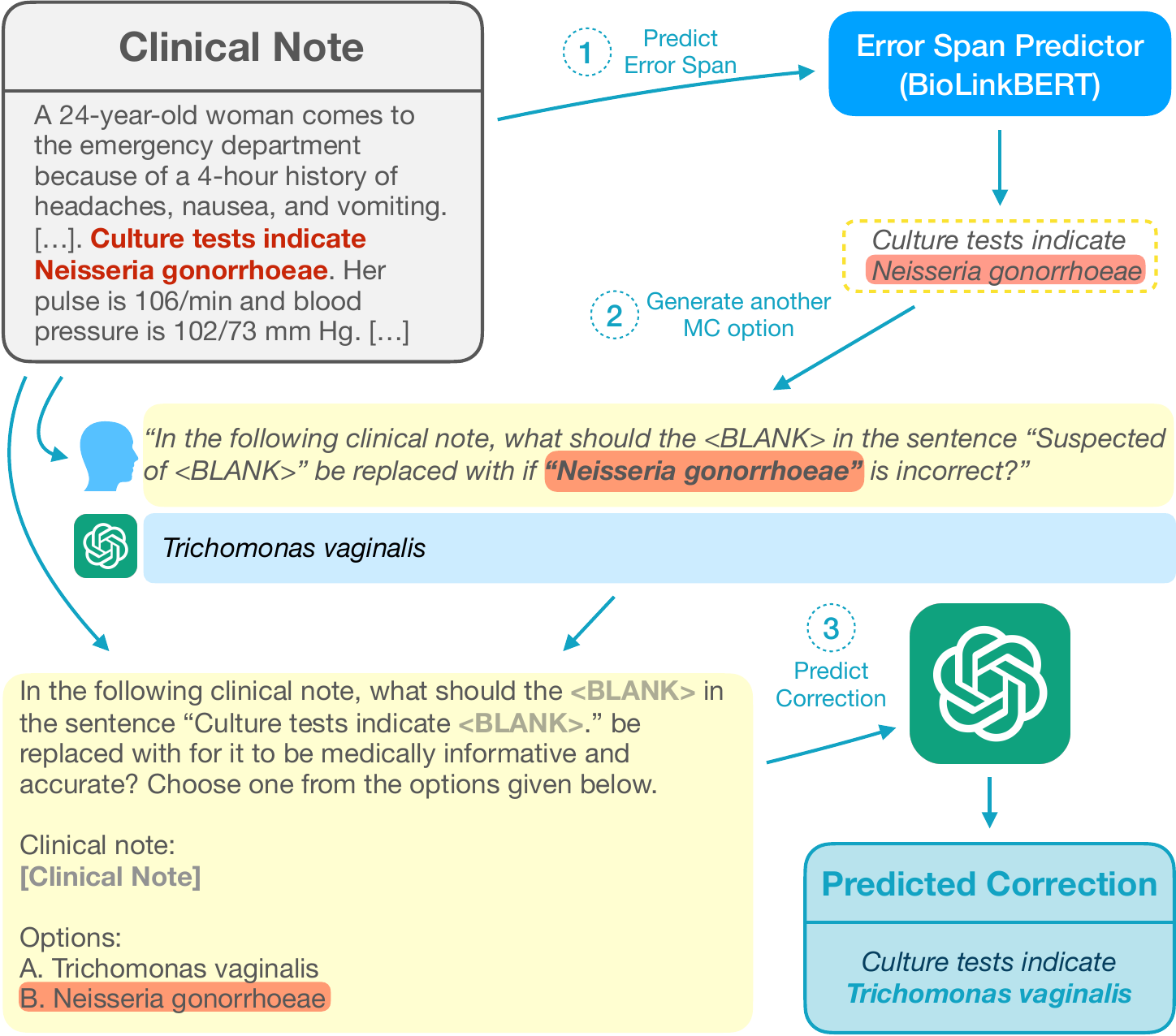}
    \caption{Schema of the Multiple-Choice Question prompt strategy.}
    \label{fig:mcq_schema}
\end{figure}

As shown in Figure~\ref{fig:mcq_schema}, this strategy involves two interactions with the LLM: 1) to construct an options set and 2) to ask a multiple-choice question.

In the first interaction, the model generates potential replacement options for the identified error span.
Here, the predicted error span is replaced with a placeholder \textit{\textcolor{gray}{``<BLANK>''}}, and the LLM is tasked with generating $n$ replacement candidates.
During our experiments, we observed a pattern where the model often included the predicted error span or its synonyms in the options.
To eliminate this redundancy, we added a directive prompt \textit{``Do not include the \textcolor{gray}{<predicted\_error\_span>} or its medical synonyms in your answer''}.

In the second interaction, we query the LLM with an MCQ-style prompt, which presents the full clinical note, with the predicted error span replaced by \textit{\textcolor{gray}{``<BLANK>''}}, and the options comprised of $n$ LLM-generated options from the first interaction and the predicted error span (totalling $n+1$ options).
The LLM chooses the best correction among these options.
Subsequently, we derive the error flag classification based on the LLM's response, \texttt{0} if it selects the predicted error span as the correct answer, or \texttt{1} if the model selects one of the other choices.
We experimented with varying the number of answer choices to two and four options.

\subsubsection{Hybrid Approach}\label{sec:hybrid_approach}

As illustrated in Figure~\ref{fig:e2e_schema}, the pipeline continues with the preparation of the ICL examples after the training for the error span prediction.
For solutions that rely only on ICL examples and do not require CoT reasoning, we directly retrieve pairs of clinical notes and their respective ground-truth corrections as ICL examples.
In contrast, CoT-based solutions require ICL examples with reasons provided.
Inspired by \citet{he2023using}, we prompted GPT-3.5 (\texttt{gpt-3.5-turbo-0613}) to generate a reasoning for the ICL examples.
We selected GPT-3.5 particularly because of its generation capability and clinical knowledge~\cite{gema2024edinburgh}.

We experimented with three CoT reasoning templates: \textbf{Brief}, \textbf{Long}, and \textbf{SOAP}.
All reasoning templates require the model to reason the ground-truth correction by identifying the incorrect span and providing the reasoning behind it.
However, each format provides a different depth and structure of reasoning.
The \textbf{Brief CoT} template prompts concise reasoning, the \textbf{Long CoT} template requires detailed step-by-step explanations, and the \textbf{SOAP CoT} template organises information according to Subjective, Objective, Assessment, and Plan sections before making corrections.

During inference, the solution uses a selected reasoning format with ICL examples to correct clinical notes.
The model applies a reasoning strategy to new scenarios based on the reasoned ICL examples which are retrieved using the BM25 algorithm~\cite{robertson1995okapi}, selecting examples similar to the clinical note in question.
We also integrate a hint about the typical nature of the errors, focusing the model's attention on specific biomedical entities such as diagnoses and treatments (\ie \textit{``Pay special attention to biomedical entities such as chief complaints, medical exams, diagnoses, and treatments.''}). We denote this as \textbf{``Type hint''}.
Finally, we leverage the error span prediction by adding it as another hint, denoted as \textbf{``Span hint''} (\ie \textit{``A clinician said that you MAY want to pay attention to the mention of \textcolor{gray}{<predicted\_error\_span>}''}).

\section{Results}

Our experiments are structured as answers to sequential research questions.
Firstly, we conducted experiments to find the best model for error span prediction, evaluating them on EM and F1 scores.
Subsequently, we experimented with various prompting strategies for error correction, evaluating them on the macro-averaged accuracy and aggregate NLG scores across MS and UW datasets.
The first error correction experiment starts with an end-to-end prompting approach, relying solely on the LLM capability with ICL and CoT to correct errors.
We, then, experimented with integrating the error span prediction model into the error correction process via the MCQ-style prompt.
Lastly, we experimented with the hybrid approach, integrating the error span prediction as a hint for the end-to-end prompting approach.
We used GPT-3.5 in our error correction experiments on the validation sets\footnote{Due to a limited research budget.}, choosing the best prompting strategy to be implemented with GPT-4 on the test set.

\subsection*{RQ1: How well are the smaller LMs performing in the error span detection?}

As shown in Table~\ref{table:res_err_span_cls}, we experimented with general (\ie BERT-base and -large) and domain-adapted models (\ie BioLinkBERT-base and -large) for the error span prediction.
We evaluated the models exclusively on a subset of the validation set that contains a medical error as stated in Subsection~\ref{sec:err_span_pred}.

Among all models, BioLinkBERT-large showed the highest EM and F1 scores on the MS validation set, indicating a superior ability to predict error spans within clinical notes.
This suggests that the domain-adaptive pretraining that BioLinkBERT has undergone contributes to its performance in medical error detection tasks.
However, all models struggle to accurately predict error spans on the UW validation set.
Recognising this, we trained BioLinkBERT-large on the MS train dataset and 25\% of the UW validation dataset as the error span prediction model for the subsequent experiments.

\begin{table}
    \centering
    \resizebox{0.95\columnwidth}{!}{
        \begin{tabular}{lcccc}
            \toprule
            \multirow{2}{*}{\textbf{Model}} & \multicolumn{2}{c}{MS} & \multicolumn{2}{c}{UW} \\
            \cmidrule(lr){2-3}
            \cmidrule(lr){4-5}
             & \textbf{EM} & \textbf{F1} & \textbf{EM} & \textbf{F1} \\
            \midrule
            BERT-base & 54.86 & 80.09 & 1.25 & 4.44 \\
            BERT-large & 55.17 & 79.30 & 5.00 & 7.92 \\
            BioLinkBERT-base & 55.17 & 81.33 & \textbf{6.25} & \textbf{12.29} \\
            BioLinkBERT-large & \textbf{58.31} & \textbf{82.49} & \textbf{6.25} & 8.91 \\
            \bottomrule
        \end{tabular}
    }
    \caption{Performance of fine-tuned error span prediction models. \textbf{Bold cell} indicates the highest score for the metric.}
    \label{table:res_err_span_cls}
\end{table}

\subsection*{RQ2: Can LLMs perform well end-to-end solely with prompting strategies?}\label{res:rq2}

Before leveraging the error span prediction, we began our error correction experiment by solely relying on the LLM with prompting strategies to correct errors without any help from the error span prediction.
This prompt-only end-to-end approach serves as the baseline for our proposed solutions.

\subsubsection*{RQ2.1: Do more ICL examples improve the LLM's performance?}

\begin{table}[t]
    \centering
        \resizebox{0.85\columnwidth}{!}{
            \begin{tabular}{cccc}
                \toprule
                \textbf{\# shots} & \textbf{$\text{Acc}_{\textbf{flag}}$} & \textbf{$\text{Acc}_{\textbf{sent\_id}}$} & \textbf{$\text{Score}_{\textbf{agg}}$} \\
                \midrule
                2 & 0.5089 & 0.3348 & 0.4139 \\
                4 & 0.5242 & 0.4215 & 0.4503 \\
                8 & \textbf{0.5268} & \textbf{0.4526} & \textbf{0.5038} \\
                \bottomrule
            \end{tabular}
        }
    \caption{Performance of GPT-3.5 using different numbers of ICL examples on validation sets. \textbf{Bold cell} indicates the highest score for the metric.}
    \label{table:res_num_icl}
\end{table}

Firstly, we experimented with varying the number of ICL examples on GPT-3.5's performance across MS and UW validation sets.
We did not report 0-shot performance as the LLM failed to generate a parseable answer, indicating that the LLM failed to complete the task without any examples.
As shown in Table~\ref{table:res_num_icl}, we observe a trend where the performance of the LLM improves in all metrics as the number of shots increases, with the 8-shot setting performing the best.
Our subsequent experiments will use the 8-shot ICL setup.

\subsubsection*{RQ2.2: Adding a hint about the typical error}

\begin{table}[t]
    \centering
    \resizebox{\columnwidth}{!}{
        \begin{tabular}{cccc}
            \toprule
            \textbf{Type Hint} & \textbf{$\text{Acc}_{\textbf{flag}}$} & \textbf{$\text{Acc}_{\textbf{sent\_id}}$} & \textbf{$\text{Score}_{\textbf{agg}}$} \\
            \midrule
            \textcolor{gray}{\xmark} & \textbf{0.5527} & 0.4472 & 0.4467 \\
            \textcolor{cyan}{\cmark} & 0.5268 \textcolor{red}{\scriptsize{(-0.03)}} & \textbf{0.4526} \textcolor{cyan}{\scriptsize{(+0.01)}} & \textbf{0.5038} \textcolor{cyan}{\scriptsize{(+0.06)}} \\
            \bottomrule
        \end{tabular}
    }
    \caption{Performance of GPT-3.5 using 8-shot prompt with or without a type hint on validation sets. Values in parentheses indicate the performance difference against the LLM that does not receive a type hint. \textcolor{cyan}{cyan} indicates improvement, \textcolor{red}{red} indicates decrease. \textbf{Bold cell} indicates the highest score for the metric.}
    \label{table:res_hint}
\end{table}

In our first experiment, we observed that the LLMs tend to correct non-essential errors (\eg grammatical and unit errors).
Thus, we prompted the LLM with a hint about the typical form of the errors (\ie \textit{``Pay special attention to biomedical entities such as chief complaints, medical exams, diagnoses, and treatments.''}).
Table~\ref{table:res_hint} shows the performance comparison between a prompt with and without this hint.
When a hint is provided, there is a decrease in the error flag accuracy by 0.03 which may indicate that there are medical errors that are not one of the specified biomedical entities.
However, this is compensated by improvements in both sentence ID accuracy and the aggregate NLG score, with the latter seeing a notable increase of 0.06.
This indicates that while the hint may slightly hinder the model's binary classification ability, it correctly directs the focus of the LLM in locating the error.

\subsubsection*{RQ2.3: Chain-of-Thought with various formats}

\begin{table}[t]
    \centering
    \resizebox{\columnwidth}{!}{
        \begin{tabular}{lccc}
            \toprule
            \textbf{CoT} & \textbf{$\text{Acc}_{\textbf{flag}}$} & \textbf{$\text{Acc}_{\textbf{sent\_id}}$} & \textbf{$\text{Score}_{\textbf{agg}}$} \\
            \midrule
            None & 0.5268 & 0.4526 & 0.5038 \\
            Brief & 0.5866 \textcolor{cyan}{\scriptsize{(+0.06)}} & \textbf{0.4989} \textcolor{cyan}{\scriptsize{(+0.05)}} & \textbf{0.5389} \textcolor{cyan}{\scriptsize{(+0.04)}} \\
            Long & \textbf{0.6074} \textcolor{cyan}{\scriptsize{(+0.08)}} & 0.4717 \textcolor{cyan}{\scriptsize{(+0.02)}} & 0.4930 \textcolor{red}{\scriptsize{(-0.01)}} \\
            SOAP & 0.5186 \textcolor{red}{\scriptsize{(-0.01)}} & 0.4058 \textcolor{red}{\scriptsize{(-0.05)}} & 0.4228 \textcolor{red}{\scriptsize{(-0.08)}} \\
            \bottomrule
        \end{tabular}
    }
    \caption{Performance of GPT-3.5 using 8-shot and type hint prompt with various CoT formats on validation sets. Values in parentheses indicate the performance difference against the LLM that does not use CoT reasoning. \textcolor{cyan}{cyan} indicates improvement, \textcolor{red}{red} indicates decrease. \textbf{Bold cell} indicates the highest score for the metric.}
    \label{table:res_cot}
\end{table}

Table \ref{table:res_cot} evaluates the effect of different Chain-of-Thought (CoT) formats on GPT-3.5's performance.
The absence of CoT (None) serves as a baseline against which the Brief, Long, and SOAP formats are compared.
The Brief CoT format leads to improvements across all metrics, particularly in sentence ID accuracy and the aggregate NLG score, underscoring the benefit of concise, targeted reasoning in enhancing model performance.
The Long format, while offering the highest accuracy in error flagging, exhibits a decrease in the aggregate score, suggesting that excessive detail may detract from overall correction quality.
Conversely, the SOAP format results in declines across all metrics, highlighting that detailed and structured reasoning approaches may not necessarily be beneficial and may even hinder the model's effectiveness.

\subsection*{RQ3: Can LLMs perform if provided with a span hint?}

After the experiments with different prompting setups, we experimented with integrating the error span prediction into the error correction process.

\subsubsection*{RQ3.1: Can LLMs perform better with MCQ-style prompts?}

As shown in Table~\ref{table:res_mcq}, MCQ-style prompt using error span prediction improved performance over end-to-end systems.
This can be attributed to two reasons.
First, the MCQ-style prompt provides options that match the specificity of the predicted error span in the original clinical note, limiting the LLMs' tendency to generate generic corrections.
Second, the MCQ-style prompt addresses the LLMs' tendency to be verbose by limiting corrections to a specific error span.


\begin{table}
    \centering
    \resizebox{\columnwidth}{!}{
        \begin{tabular}{lccc}
            \toprule
            \textbf{Prompting Strategy} & \textbf{$\text{Acc}_{\textbf{flag}}$} & \textbf{$\text{Acc}_{\textbf{sent\_id}}$} & \textbf{$\text{Score}_{\textbf{agg}}$} \\
            \midrule
            8-shot + Brief CoT & 0.5866 & 0.4989 & 0.5389 \\
            \midrule
            MCQ (2 options) & \textbf{0.6131} & \textbf{0.6029} & \textbf{0.6492} \\
            MCQ (4 options) & 0.6087 & 0.5944 & 0.6448 \\
            \bottomrule
        \end{tabular}
    }
    \caption{Performance of GPT-3.5 with the MCQ-style prompt on validation sets. \textbf{Bold cell} indicates the highest score for the metric.}
    \label{table:res_mcq}
\end{table}

\subsubsection*{RQ3.2: Can end-to-end LLMs perform better when provided with a span hint?}


In our RQ2 experiments with end-to-end systems, we observed limitations in the LLM's ability to accurately locate errors within the clinical notes.
While in RQ3.1, we noticed that integrating error span predictions helped improve the LLM's performance.
These insights motivated us to integrate the error span predictions from fine-tuned models to the end-to-end LLM solution.
We denoted this solution as the ``Hybrid approach'', as mentioned in Subsubsection~\ref{sec:hybrid_approach}, leveraging the ``Span hint'' from the error span prediction. 

\begin{table}
    \centering
    \resizebox{\columnwidth}{!}{
        \begin{tabular}{lcccc}
            \toprule
            \textbf{CoT} & \textbf{Span Hint} & \textbf{$\text{Acc}_{\textbf{flag}}$} & \textbf{$\text{Acc}_{\textbf{sent\_id}}$} & \textbf{$\text{Score}_{\textbf{agg}}$} \\
            \midrule
            MCQ (2 opt) & \textcolor{cyan}{\cmark} & \textbf{0.6131} & \textbf{0.6029} & 0.6492 \\
            MCQ (4 opt) & \textcolor{cyan}{\cmark} & 0.6087 & 0.5944 & 0.6448 \\
            \midrule
            \multirow{2}{*}{None}  & \textcolor{gray}{\xmark} & 0.5268 & 0.4526 & 0.5038 \\
                                   & \textcolor{cyan}{\cmark} & 0.5671 \textcolor{cyan}{\scriptsize{(+0.04)}} & 0.5543 \textcolor{cyan}{\scriptsize{(+0.10)}} & 0.7348 \textcolor{cyan}{\scriptsize{(+0.23)}} \\
            \midrule
            \multirow{2}{*}{Brief} & \textcolor{gray}{\xmark} & 0.5866 & 0.4989 & 0.5389 \\
                                   & \textcolor{cyan}{\cmark} & 0.5610 \textcolor{red}{\scriptsize{(-0.03)}} & 0.5454 \textcolor{cyan}{\scriptsize{(+0.05)}} & \textbf{0.7385} \textcolor{cyan}{\scriptsize{(+0.20)}} \\
            \midrule
            \multirow{2}{*}{Long}  & \textcolor{gray}{\xmark} & 0.6074 & 0.4717 & 0.4930 \\
                                   & \textcolor{cyan}{\cmark} & 0.6048 \textcolor{red}{\scriptsize{(-0.00)}} & 0.4651 \textcolor{red}{\scriptsize{(-0.01)}} & 0.4822 \textcolor{red}{\scriptsize{(-0.01)}} \\
            \midrule
            \multirow{2}{*}{SOAP}  & \textcolor{gray}{\xmark} & 0.5186 & 0.4058 & 0.4228 \\
                                   & \textcolor{cyan}{\cmark} & 0.5237 \textcolor{cyan}{\scriptsize{(+0.01)}} & 0.4310 \textcolor{cyan}{\scriptsize{(+0.03)}} & 0.4884 \textcolor{cyan}{\scriptsize{(+0.07)}} \\
            \bottomrule
        \end{tabular}
    }
    \caption{Performance of GPT-3.5 using 8-shot and type hint prompt with various CoT format and with or without receiving span hint on validation sets. Values in parentheses indicate the performance difference against the solution that does not receive a span hint. \textcolor{cyan}{cyan} indicates improvement, \textcolor{red}{red} indicates decrease. \textbf{Bold cell} indicates the highest score for the metric.}
    \label{table:res_cot_span_hint}
\end{table}

Integrating a span hint into the end-to-end LLM prompt resulted in improvements across all metrics, as shown in Table~\ref{table:res_cot_span_hint}.
Notably, span hint significantly improved the aggregate NLG scores of Brief CoT and no-CoT solutions.
However, span hint did not improve Long CoT solution, suggesting that the reasoning style may influence the LLM's ability to leverage span hints.

Despite MCQ prompts demonstrating higher accuracy in error sentence identification, ``Brief CoT'' prompts combined with ICL, type hint, and span hints showed a higher aggregate NLG score, emphasising the different strengths of the two strategies.
This indicates that the hybrid approach harnesses the LLM's generative capabilities, while the fine-tuned error span prediction model helps direct these corrections to the appropriate error locations.




\subsection*{Performance on Test Set}

\begin{table}
    \centering
    \resizebox{\columnwidth}{!}{
        \begin{tabular}{lccc}
            \toprule
            \textbf{Prompting Strategy} & \textbf{$\text{Acc}_{\textbf{flag}}$} & \textbf{$\text{Acc}_{\textbf{sent\_id}}$} & \textbf{$\text{Score}_{\textbf{agg}}$} \\
            \midrule
            8-shot + Hints & 0.5243 & 0.4649 & 0.6274 \\
            \textbf{8-shot + Brief CoT + Hints} & \textbf{0.6681} & 0.5924 & \textbf{0.6634} \\
            \midrule
            MCQ (2 options) & 0.6573 & \textbf{0.5957} & 0.6267 \\
            MCQ (4 options) & 0.5935 & 0.5232 & 0.5882 \\
            \bottomrule
        \end{tabular}
    }
    \caption{Results of GPT-4 with either ICL + CoT + hinted prompt or Multiple-Choice-Question prompt on test sets. The models are compared based on the aggregate NLG score.}
    \label{table:res_test}
\end{table}


We submitted our four best-performing solutions to be evaluated on the holdout test set.
As shown in Table~\ref{table:res_test}, we can observe a similar trend as in the validation set experiments.
The 2-options MCQ prompts show strong performance in accurately identifying the error-containing sentence.
The 8-shot + Brief CoT + Hints method performs better, especially in the aggregate NLG score.
This suggests that while MCQ prompts effectively direct the model's focus, enabling accurate detection of errors, they may slightly constrain the model's generative capability.
Overall, these results highlight the benefit of using concise CoT reasoning in LLMs as well as providing guidance via targeted hints.
Our best-performing pipeline, 8-shot + Brief CoT + Hints, ranked sixth in the shared task leaderboard based on the aggregate NLG score.

\section{Post-hoc Analyses}

Commonly reported NLG metrics tend to not be well correlated with human judgement, especially in the clinical domain~\cite{ben-abacha-etal-2023-investigation}.
To understand the limitations of LLMs for clinical note correction, we extend beyond the reported performance metrics by analysing the sensitivity of LLMs to the data and prompt, as well as the common mistakes that LLMs tend to commit.\footnote{Post-hoc analyses are conducted on the validation sets.}

\subsection{Sensitivity}

It is a well-known fact that the performance of an LLM may differ massively given slight differences in the way we prompt it~\cite{voronov2024mind}.
We analysed factors observed in the data and prompt that may contribute to performance differences.

\subsubsection{Sensitivity to the position of error sentence in the clinical note}

\begin{figure*}
    \centering
    \includegraphics[width=\textwidth]{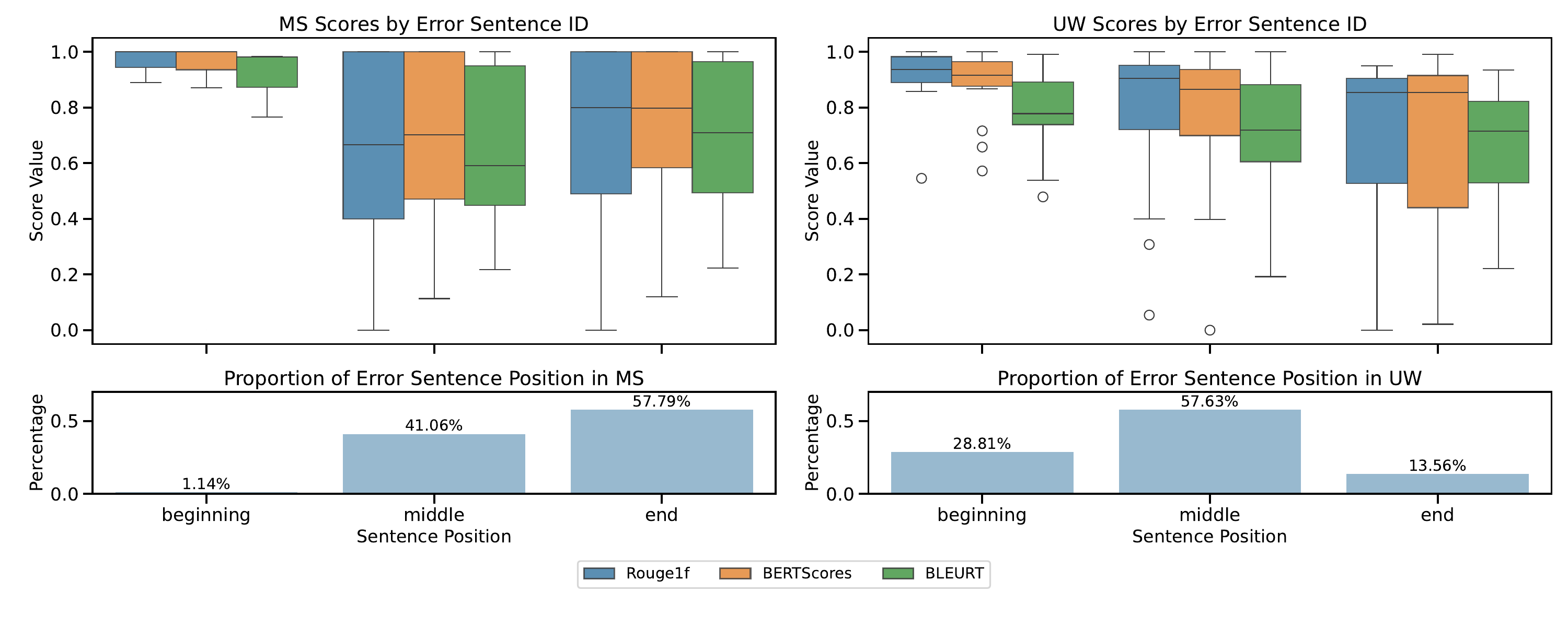}
    \caption{Boxplots of the distribution of ROUGE 1, BERTScore, and BLEURT with respect to the position of the error sentence for MS (left) and UW (right) datasets. ``beginning'' denotes that the error sentence is at index 0, ``end'' at the end, while ``middle'' is in between ``beginning'' and ``end''.}
    \label{fig:res_sens_sent_id}
\end{figure*}

We investigated the sensitivity of the model performance to the position of the error sentence within a given clinical note, dividing them into three cases; if the error sentence is in the first sentence (\textit{``beginning''}), the last sentence (\textit{``end''}), or in between the first and the last sentences (\textit{``middle''}).

Figure~\ref{fig:res_sens_sent_id} illustrates the relationship between the NLG metrics and the error sentence position, along with the proportion of the error sentence location.
We can observe that ROUGE 1, BERTScore, and BLEURT scores do not vary significantly based on the position of the error sentence.
This observation is quantitatively supported by the Kruskal-Wallis H-Test and the post-hoc Dunn's test results shown in Appendix~\ref{app:sens_stats}.
The test results reveal that the LLM's ability to generate accurate corrections is not impacted by where the error appears in the input, which is a desirable trait.

\subsubsection{Sensitivity to the role described in the system prompt}

\begin{table}
    \centering
    \resizebox{\columnwidth}{!}{
        \begin{tabular}{lccc}
            \toprule
            \textbf{Role} & \textbf{$\text{Acc}_{\textbf{flag}}$} & \textbf{$\text{Acc}_{\textbf{sent\_id}}$} & \textbf{$\text{Score}_{\textbf{agg}}$} \\
            \midrule
            Clinician assistant & 0.5610 & 0.5454 & 0.7385 \\
            \midrule
            No role & 0.5570 \textcolor{red}{\scriptsize{(-0.00)}} & 0.5416 \textcolor{red}{\scriptsize{(-0.00)}} & 0.7504 \textcolor{cyan}{\scriptsize{(+0.01)}} \\
            Assistant & 0.5509 \textcolor{red}{\scriptsize{(-0.01)}} & 0.5442 \textcolor{red}{\scriptsize{(-0.00)}} & 0.7504 \textcolor{cyan}{\scriptsize{(+0.01)}} \\
            Medical student & 0.5539 \textcolor{red}{\scriptsize{(-0.01)}} & 0.5468 \textcolor{cyan}{\scriptsize{(+0.00)}} & 0.7484 \textcolor{cyan}{\scriptsize{(+0.01)}} \\
            Nurse & 0.5763 \textcolor{cyan}{\scriptsize{(+0.02)}} & 0.5615 \textcolor{cyan}{\scriptsize{(+0.02)}} & 0.7424 \textcolor{cyan}{\scriptsize{(+0.00)}} \\
            Clinical note verificator & 0.5554 \textcolor{cyan}{\scriptsize{(+0.01)}} & 0.5438 \textcolor{red}{\scriptsize{(-0.00)}} & 0.7518 \textcolor{cyan}{\scriptsize{(+0.01)}} \\
            Clinician & 0.5793 \textcolor{cyan}{\scriptsize{(+0.02)}} & 0.5615 \textcolor{cyan}{\scriptsize{(+0.02)}} & 0.7615 \textcolor{cyan}{\scriptsize{(+0.02)}} \\
            \bottomrule
        \end{tabular}
    }
    \caption{Performance of our best-performing solution when prompted with different roles via the system prompt (\ie \textit{``You are <<a role>> tasked to ...''}) on the validation sets.}
    \label{table:res_roleplay}
\end{table}

Owing to their instruction-following ability, LLMs are capable of playing a role as prompted by the user~\cite{wang2023aligning}. 
In the clinical domain, we tend to prompt an LLM to answer a query as a healthcare professional, such as a clinician.
In this analysis, we explored how the role prompted or the lack thereof may affect the performance of the LLM in generating corrections.
We modify the system prompt (\ie \textit{``You are <<a role>> tasked to ...''}) with various role options.
Table~\ref{table:res_roleplay} details the varying performances of the best-performing 8-shot + Brief CoT + hints solution when prompted with different roles.
The LLM performs best when prompted to role-play as a ``clinician''.
This phenomenon, known as \emph{In-Context Impersonation}~\cite{salewski2024context}, highlights that role-playing should be examined when developing a prompt-based solution.

\subsubsection{Sensitivity to the position of the multiple choice options}

\begin{table}
    \centering
    \resizebox{\columnwidth}{!}{
        \begin{tabular}{cccc}
            \toprule
            \textbf{Generated Option Position} & \textbf{$\text{Acc}_{\textbf{flag}}$} & \textbf{$\text{Acc}_{\textbf{sent\_id}}$} & \textbf{$\text{Score}_{\textbf{agg}}$} \\
            \midrule
            A & 0.6131 & 0.6029 & \textbf{0.6492} \\
            B & \textbf{0.6368} & \textbf{0.6265} & 0.6380 \\
            \bottomrule
        \end{tabular}
    }
    \caption{Results of the sensitivity analysis of MCQ-style prompt to the position of the LLM-generated option in the 2 options setting on validation sets.}
    \label{table:res_sens_mcq}
\end{table}

Table~\ref{table:res_sens_mcq} shows the outcome of a sensitivity analysis, based on the relative positioning of the LLM-generated option and the predicted error span within the original text for the systems with MCQ-type prompts.
Both binary classification accuracy and error sentence prediction accuracy were improved when the LLM-generated option was positioned as option B, as opposed to option A.
On the other hand, the aggregate score for correction reveals a higher score when the LLM-generated option was positioned as option A, achieving a score of 0.6492.
This observation of \textit{selection bias} echoes findings by previous studies~\cite{pezeshkpour2023large,zheng2023large}.

\subsection{Common LLM mistakes}




We qualitatively evaluated the common mistakes found in the generated reasons and corrections.

\paragraph{Corrections of marginal effects}
LLMs occasionally make minor corrections to clinical notes that, although technically correct, do not significantly affect the correctness.
Changes, such as altering ``3'' to ``three'' or fixing grammatical mistakes, might enhance readability but are not clinically significant.
LLMs also tend to add adjectives, such as ``acute'' to ``pyelonephritis'', adding specificity desirable in clinical settings but not always favourably reflected in NLG metrics.

\paragraph{Near-accurate corrections}


LLMs often suggest near-accurate corrections that lack the required specificity.
For example, fixing an error sentence with the generic ``antiplatelet therapy'' instead of ``aspirin'' misses the required precision, even though aspirin is an antiplatelet therapy.
Likewise, proposing to ``Start anticoagulation therapy'' instead of the more explicit ``dalteparin'' lacks specificity.
These near-accurate adjustments underscore the difficulty LLMs encounter in achieving the specificity of the ground truth label.

\paragraph{Mistake due to incomplete context}


LLMs struggle to fix errors in clinical notes when details are lacking.
One example is when the LLM mistakenly suggests changing ``pulmonary fibrosis'' to ``chronic obstructive pulmonary disease''.
Both conditions share very similar early symptoms that are difficult to differentiate even for clinicians~\cite{chilosi2012pathogenesis}.
Another example involves incorrectly adjusting a malnutrition patient's Body Mass Index (BMI) from 30 to 18.
Albeit a BMI of 18 signals malnutrition, it deviates from the ground truth label 13.
These instances underscore the complexity of the MEDIQA-CORR task, as well as medical error correction in general which is very challenging to do without additional context even for human clinicians.

In summary, the sensitivity and qualitative analyses highlight the current limitations of LLMs in the clinical domain, which prompt further questions about the readiness of LLMs to be implemented in real-world clinical settings.

\section{Conclusion}


This study explores strategies for using LLMs to detect and correct medical error for the MEDIQA-CORR 2024 shared task.
In addition to the comprehensive evaluation of prompting strategies based on different reasoning styles, we experiment with integrating error-span predictions from a fine-tuned model.
Our best-performing system includes a fine-tuned BioLinkBERT-large for error-span prediction and GPT-4 for error correction. By harnessing LLMs' generative abilities with 8-shot ICL and Brief CoT and presenting predicted error span as a hint in the prompt, our best-performing solution ranked sixth in the shared task leaderboard.
Our post-hoc analyses offer insights into the use of LLM in medical error correction, including sensitivity to error location, role-playing bias, and common types of mistakes made by LLMs.


\section*{Limitations}


The scope of our study was exclusively confined to GPT-based models, namely GPT-3.5 and GPT-4.
The reported findings may differ across different types of LLMs.
Furthermore, we independently explored various prompting strategies, such as CoT and MCQ prompt.
We did not investigate the effect of integrating MCQ prompt with CoT reasoning.
This unexplored combination may offer additional improvements in the LLM's error correction capabilities.

Our post-hoc analyses also reveal a significant limitation of LLMs in clinical settings.
Despite the advancements demonstrated through our proposed methodologies, the study underscores that LLMs may not be ready for deployment in real-world clinical environments without human oversight.
The analysis highlights the critical need for human supervision, especially given the potential risks associated with inaccuracies in medical documentation and the consequent impacts on patient care.
This limitation calls for further research into enhancing the reliability of LLMs as well as the evaluation metrics before considering their implementation in sensitive areas such as healthcare.

\section*{Acknowledgements}

APG and CL were supported by the United Kingdom Research and Innovation (grant EP/S02431X/1), UKRI Centre for Doctoral Training in Biomedical AI at the University of Edinburgh, School of Informatics. 
PM was partially funded by 
ELIAI (The Edinburgh Laboratory for Integrated Artificial Intelligence), EPSRC (grant no. EP/W002876/1), an industry grant from Cisco, and a donation from Accenture LLP; and is grateful to NVIDIA for the GPU donations.
BA was partially funded by Legal and General PLC as part of the Advanced Care Research Centre and by the Artificial Intelligence and Multimorbidity: Clustering in Individuals, Space and Clinical Context (AIM-CISC) grant NIHR202639.
For the purpose of open access, The authors have applied a Creative Commons attribution (CC BY) licence to any author-accepted manuscript version arising.
Experiments from this work are conducted mainly on the Edinburgh International Data Facility\footnote{https://edinburgh-international-data-facility.ed.ac.uk/} and supported by the Data-Driven Innovation Programme at the University of Edinburgh.

\bibliography{anthology,custom}
\bibliographystyle{acl_natbib}

\appendix

\section{Experimental setup}
\label{sec:app_setup}
All fine-tuning experiments were run on a single NVIDIA A100-40GB GPUs.
We used the HuggingFace's transformer library~\cite{wolf-etal-2020-transformers}.
The validation set was utilised to determine the best checkpoint.

In-context examples were retrieved from the Training set.
Additionally, the validation set was used to evaluate and select the optimal prompt design.
For the test submission, we also retrieved In-context examples from the MS and UW validation sets.

\section{Hyperparameters}
\label{sec:app_hyperparams}

\subsection{GPT-3.5 Hyperparameters for the generation of Natural Language Explanation}
\label{sec:chatgpt_hyperparams}

We prompted GPT-3.5 (model name: \texttt{gpt-3.5-turbo-0613}) with hyperparameters as shown in Table~\ref{tab:api_hyperparameters}.
The generation process took approximately 2 hours and cost \$2.

\begin{table}
    \centering
    \small
    \begin{tabular}{ll}
    \toprule
    {\bf Parameter} & {\bf Value} \\
    \midrule
    Model Name        & gpt-3.5-turbo-0613 \\
    API Version       & 2023-03-15-preview \\
    Temperature       & 0 \\
    Top P             & 0 \\
    Frequency Penalty & 0 \\
    Presence Penalty  & 0 \\
    Max new token     & 256 \\
    \bottomrule
    \end{tabular}
    \caption{GPT-3.5 API call hyperparameters to generate Natural Language Explanations.}
    \label{tab:api_hyperparameters}
\end{table}

\subsection{GPT-4 generation hyperparameters}
\label{sec:gpt4_gen_hyperparams}

During inference on the test set, we prompted GPT-4 (model name: \texttt{gpt-4-turbo}) as shown in Figure~\ref{fig:e2e_schema} Step 3.
We set \texttt{temperature=0} to ensure that the model's generation is deterministic.
The maximum generation length is 512, allowing longer CoT reasons.
One generation process took approximately 2 hours and cost \$35.

\section{Prompt Examples}
\label{sec:prompt_examples}
Here, we provide examples of the prompts used in our experiments. The black text within the box represents the prompt input text, the \textcolor{red}{red} text represents the prediction of the models, and the \textcolor{blue}{blue} text represents the ground truth.

\subsection{Prompt for In-Context Learning and Chain-of-Thought}
\label{sec:example_icl}

\subsubsection{System Prompt}

\begin{tcolorbox}[colback=sepia!25!white,colframe=sepia!100!sepia,colbacktitle=sepia!90!white, title=System Prompt for the In-Context Learning and Chain-of-Thought Experiments,fonttitle=\bfseries]
\small
You are a clinician assistant tasked with reviewing clinical texts that have been corrupted by an evil third party. Each clinical text may contain either one perturbed sentence with clinical or factual inaccuracies, or no errors at all. Your objective is to help the clinician in identifying any perturbed sentence and provide a correction. Please respond in JSON format with the following structure:\\\\- `reason': Describe the reasoning behind identifying a specific sentence as incorrect or all clinical text as correct.\\- `incorrect\_sentence\_id': If you identify an incorrect sentence, provide its ID here. If all sentences are correct, respond with -1.\\- 'correction': If an incorrect sentence is identified, provide a corrected sentence or an explanation of the inaccuracy. If all sentences are correct, respond with 'NA'.\\\\When evaluating the text, focus specifically on clinical or factual inaccuracies. This could include incorrect medical information, factual errors related to patient care, or erroneous data interpretations. Your detailed assessment and correction are critical for ensuring the reliability and accuracy of our clinical documentation. You have to be more action-oriented, don't suggest planning, discussion, or something that is not action-oriented.
\end{tcolorbox}

\subsubsection{CoT reasons}

\begin{tcolorbox}[colback=blue!25!white,colframe=blue!100!blue,colbacktitle=blue!90!white, title=Brief Chain-of-Thought,fonttitle=\bfseries]
\small
Streptococcus pneumoniae is not typically associated with skin lesions with a blue livid center, pale intermediate zone, and a dark red peripheral rim. Mycoplasma pneumoniae is a common cause of atypical pneumonia and can present with skin rashes like the ones described in the text.
\end{tcolorbox}

\begin{tcolorbox}[colback=blue!25!white,colframe=blue!100!blue,colbacktitle=blue!90!white, title=Long Chain-of-Thought,fonttitle=\bfseries]
\small
The incorrect sentence is sentence 24, which states `Pathology confirms Streptococcus pneumoniae is the causal organism.' This is inaccurate because Streptococcus pneumoniae is a common cause of community-acquired pneumonia, but in this case, the clinical presentation with skin lesions and the absence of a productive cough suggests a more likely diagnosis of atypical pneumonia caused by Mycoplasma pneumoniae. Mycoplasma pneumoniae is known to cause atypical pneumonia with extrapulmonary manifestations such as skin lesions, which aligns better with the patient's presentation. Therefore, the correction should be `Pathology confirms Mycoplasma pneumoniae is the causal organism.'
\end{tcolorbox}

\begin{tcolorbox}[colback=blue!25!white,colframe=blue!100!blue,colbacktitle=blue!90!white, title=SOAP Chain-of-Thought,fonttitle=\bfseries]
\small
Subjective: The patient presents with fever, nausea, nonproductive cough, headaches, generalized fatigue, muscle and joint pain, and increasing shortness of breath.\\
Objective: Vital signs show a slightly elevated temperature, normal pulse, respirations, and blood pressure. Skin lesions with specific characteristics are noted on the extremities. Laboratory studies reveal hemoglobin, leukocyte count, electrolyte levels, and other values. An x-ray of the chest is performed.\\
Assessment: The patient likely has an infectious process given the symptoms and laboratory findings.\\
Plan: Further evaluation and treatment are warranted.\\
Inconsistency: The incorrect sentence is in the Assessment section where Streptococcus pneumoniae is mentioned as the causal organism, which is inconsistent with the typical causative agent for atypical pneumonia, which is Mycoplasma pneumoniae.
\end{tcolorbox}

\newpage

\subsubsection{Chain-of-Thought Prompt}
\begin{tcolorbox}[colback=cyan!25!white,colframe=cyan!100!cyan,colbacktitle=cyan!90!white, title=Brief Chain-of-Thought Prompt Example,fonttitle=\bfseries]
\small
\textbf{\textcolor{sepia}{<<System Prompt>>}}\\
\textbf{\textcolor{blue}{<<In-Context Learning Examples>>}}\\
Clinical text:\\\\
\textcolor{gray}{0 A 78-year-old man is brought in to the emergency department by ambulance after his wife noticed that he began slurring his speech and had developed facial asymmetry during dinner approximately 30 minutes ago. \\
1 His past medical history is remarkable for hypertension and diabetes. \\
2 His temperature is 99.1 F (37.3 C), blood pressure is 154/99 \\
3 mmHg, pulse is 89/min, respirations are 12/min, and oxygen saturation is 98\% on room air. \\
4 CTA of the head is obtained after neurologic exam reveals right upper and lower extremity weakness and an asymmetric smile.}\\
\\
Task: Identify 1 incorrect sentence in the clinical text, or confirm if all sentences are correct. If there is 1 incorrect sentence, how would you fix it? Follow the hints below if you see fit\\
Hint:\\
- Pay special attention to biomedical entities such as chief complaints, medical exams, diagnoses, and treatments. The mistake often lies within these areas.\\
- The clinician said that you MAY want to pay attention to the mention of '\textbf{\textcolor{cyan}{CTA of the head}}'. If you believe that the mention is incorrect, ONLY SWAP this mention with something more probable. DO NOT MODIFY the sentence in any other way.\\
Let's think step by step\\
Answer: \\
\textbf{\textcolor{cyan}{\{}}\\
\hspace{5em}\textcolor{cyan}{\textbf{``reason''}: ``The mention of `CTA of the head' is not the most appropriate imaging study for the patient's presentation of acute stroke symptoms with slurred speech, facial asymmetry, and extremity weakness. A non-contrast CT of the head is typically the initial imaging study to evaluate for acute intracranial haemorrhage or ischemic stroke. Therefore, replacing `CTA of the head' with `CT of the head' would align better with the standard diagnostic approach for acute stroke evaluation.'',} \\
\hspace{5em}\textcolor{cyan}{\textbf{``incorrect\_sentence\_id''}: ``4'',}\\
\hspace{5em}\textcolor{cyan}{\textbf{``correction''}: ``CT of the head is obtained after neurologic exam reveals right upper and lower extremity weakness and an asymmetric smile.'',}\\
\textbf{\textcolor{cyan}{\}}}
\end{tcolorbox}

\newpage

\subsection{Option Generation Prompt Multiple-Choice Question Prompt}
\label{sec:example_mcq_opt_gen}

\subsubsection{MCQ 2 options}

\begin{tcolorbox}[colback=cyan!25!white,colframe=cyan!100!cyan,colbacktitle=cyan!90!white, title=Prompt Used to Generate MCQ Answer Options,fonttitle=\bfseries]
\small
Your job is to review a clinical note that potentially contains a medical error.\\

In the following clinical note, what should the \textcolor{gray}{<BLANK>} in the sentence ``Suspected of \textcolor{gray}{<BLANK>}.'' be replaced with if "primary ciliary dyskinesia" is incorrect? Do not answer with "primary ciliary dyskinesia" or its medical synonyms in your answer. Output your response in JSON format, with keys 'option'.
\\
\\
Clinical note:
\\
\\
\textcolor{gray}{A 4-year-old boy is brought to the physician in December for episodic shortness of breath and a nonproductive cough for 3 months. These episodes frequently occur before sleeping, and he occasionally wakes up because of difficulty breathing. His mother also reports that he became short of breath while playing with his friends at daycare on several occasions. He is allergic to peanuts. He is at the 55th percentile for height and weight. Vital signs are within normal limits. Examination shows mild scattered wheezing in the thorax. An x-ray of the chest shows no abnormalities. Suspected of <BLANK>.}
\\
\\
Generated answer:\\
\textcolor{cyan}{\{\\
\textbf{"option"}: "asthma"\\
\}}

\end{tcolorbox}

\begin{tcolorbox}[colback=white!0!white,colframe=white!0!white,colbacktitle=white!0!white, title= ,fonttitle=\bfseries]
\end{tcolorbox}
\begin{tcolorbox}[colback=white!0!white,colframe=white!0!white,colbacktitle=white!0!white, title= ,fonttitle=\bfseries]
\end{tcolorbox}
\begin{tcolorbox}[colback=white!0!white,colframe=white!0!white,colbacktitle=white!0!white, title= ,fonttitle=\bfseries]
\end{tcolorbox}
\begin{tcolorbox}[colback=white!0!white,colframe=white!0!white,colbacktitle=white!0!white, title= ,fonttitle=\bfseries]
\end{tcolorbox}
\begin{tcolorbox}[colback=white!0!white,colframe=white!0!white,colbacktitle=white!0!white, title= ,fonttitle=\bfseries]
\end{tcolorbox}

\newpage

\subsubsection{MCQ 4 options}

\begin{tcolorbox}[colback=cyan!25!white,colframe=cyan!100!cyan,colbacktitle=cyan!90!white, title=Prompt Used to Generate MCQ Answer Options,fonttitle=\bfseries]
\small
Your job is to review a clinical note that potentially contains a medical error.\\


In the following clinical note, what should the \textcolor{gray}{<BLANK>} in the sentence ``Suspected of \textcolor{gray}{<BLANK>}.'' be replaced with if "primary ciliary dyskinesia" is incorrect? Do not answer with "primary ciliary dyskinesia" or its medical synonyms in your answer. Output your response in JSON format, with keys 'option\_1', 'option\_2' and 'option\_3'.
\\
\\
Clinical note:
\\
\\
\textcolor{gray}{A 4-year-old boy is brought to the physician in December for episodic shortness of breath and a nonproductive cough for 3 months. These episodes frequently occur before sleeping, and he occasionally wakes up because of difficulty breathing. His mother also reports that he became short of breath while playing with his friends at daycare on several occasions. He is allergic to peanuts. He is at the 55th percentile for height and weight. Vital signs are within normal limits. Examination shows mild scattered wheezing in the thorax. An x-ray of the chest shows no abnormalities. Suspected of <BLANK>.}
\\
\\
Generated answer:\\
\textcolor{cyan}{\{\\
\textbf{"option\_1"}: "asthma"\\
\textbf{"option\_2"}: "bronchiolitis"\\
\textbf{"option\_3"}: "pulmonary embolism"\\
\}}
\end{tcolorbox}

\newpage

\subsection{Inference Prompt Multiple-Choice Question Prompt}
\label{sec:example_mcq_inference}

\subsubsection{MCQ 2 options}

\begin{tcolorbox}[colback=cyan!25!white,colframe=cyan!100!cyan,colbacktitle=cyan!90!white, title=Inference Prompt for Multiple-Choice Question style with 2 options,fonttitle=\bfseries]
\small
Your job is to review a clinical note that potentially contains a medical error.\\


In the following clinical note, what should the \textcolor{gray}{<BLANK>} in the sentence ``Suspected of \textcolor{gray}{<BLANK>}.'' be replaced with for it to be medically informative and accurate? Choose one from the options given below. Output your response in JSON format, with a key `Answer'.
\\
\\
Clinical note:
\\
\\
\textcolor{gray}{A 4-year-old boy is brought to the physician in December for episodic shortness of breath and a nonproductive cough for 3 months. These episodes frequently occur before sleeping, and he occasionally wakes up because of difficulty breathing. His mother also reports that he became short of breath while playing with his friends at daycare on several occasions. He is allergic to peanuts. He is at the 55th percentile for height and weight. Vital signs are within normal limits. Examination shows mild scattered wheezing in the thorax. An x-ray of the chest shows no abnormalities. Suspected of <BLANK>.}
\\
\\
Options:
\\
\\
	\indent A. asthma \\
	\qquad B. primary ciliary dyskinesia \\

Generated answer:
\textcolor{cyan}{\{\\
\textbf{"Answer"}: "A. asthma"\\
\}}

\end{tcolorbox}

\newpage

\subsubsection{MCQ 4 options}

\begin{tcolorbox}[colback=cyan!25!white,colframe=cyan!100!cyan,colbacktitle=cyan!90!white, title=Inference Prompt for Multiple-Choice Question style with 4 options,fonttitle=\bfseries]
\small
Your job is to review a clinical note that potentially contains a medical error.\\


In the following clinical note, what should the \textcolor{gray}{<BLANK>} in the sentence ``Culture tests indicate \textcolor{gray}{<BLANK>}.'' be replaced with for it to be medically informative and accurate? Choose one from the options given below. Output your response in JSON format, with a key `Answer'.
\\
\\
Clinical note:
\\
\\
\textcolor{gray}{A 4-year-old boy is brought to the physician in December for episodic shortness of breath and a nonproductive cough for 3 months. These episodes frequently occur before sleeping, and he occasionally wakes up because of difficulty breathing. His mother also reports that he became short of breath while playing with his friends at daycare on several occasions. He is allergic to peanuts. He is at the 55th percentile for height and weight. Vital signs are within normal limits. Examination shows mild scattered wheezing in the thorax. An x-ray of the chest shows no abnormalities. Suspected of <BLANK>.}
\\
\\
Options:
\\
\\
	\indent A. asthma \\
	\qquad B. primary ciliary dyskinesia \\
	\qquad C. bronchiolitis \\
	\qquad D. pulmonary embolism \\

Generated answer:
\textcolor{cyan}{\{\\
\textbf{"Answer"}: "A. asthma"\\
\}}

\end{tcolorbox}

\newpage

\section{Statistics of ``Sensitivity to the position of error sentence in the clinical note''}
\label{app:sens_stats}

\begin{table}
    \centering
    \resizebox{\columnwidth}{!}{
        \begin{tabular}{lcccccc}
            \toprule
            & \multicolumn{3}{c}{\textbf{MS}} & \multicolumn{3}{c}{\textbf{UW}} \\
            \cmidrule(lr){2-4}
            \cmidrule(lr){5-7}
            & \textbf{ROUGE 1} & \textbf{BERTScore} & \textbf{BLEURT} & \textbf{ROUGE 1} & \textbf{BERTScore} & \textbf{BLEURT} \\
            \midrule
            $H$ & 6.0749 & 5.0249 & 7.2848 & 5.6821 & 3.6073 & 2.3457 \\
            $p$ & \textcolor{cyan}{\textbf{0.0480}} & 0.0811 & \textcolor{cyan}{\textbf{0.0262}} & 0.0584 & 0.1647 & 0.3095 \\
            \bottomrule
        \end{tabular}
    }
    \caption{Summary of Kruskal-Wallis H-Test results for sentence position impact on ROUGE 1, BERTScore, and BLEURT metrics. Statistically significant differences (p < 0.05) are highlighted in \textcolor{cyan}{\textbf{cyan}}.}
    \label{tab:res_sens_sent_id}
\end{table}

\begin{table}
    \centering
    \resizebox{\columnwidth}{!}{
        \begin{tabular}{lcccccc}
            \toprule
            & \multicolumn{3}{c}{\textbf{MS}} & \multicolumn{3}{c}{\textbf{UW}} \\
            \cmidrule(lr){2-4}
            \cmidrule(lr){5-7}
            & \textbf{ROUGE 1} & \textbf{BERTScore} & \textbf{BLEURT} & \textbf{ROUGE 1} & \textbf{BERTScore} & \textbf{BLEURT} \\
            \midrule
            beginning-middle & 0.1751 & 0.3121 & 0.1389 & 0.3596 & 0.3464 & 0.7118 \\
            middle-end & 0.2137 & 0.2479 & 0.1192 & 0.5251 & 1.0000 & 1.0000 \\
            beginning-end & 0.3923 & 0.6258 & 0.3586 & 0.0609 & 0.2849 & 0.4757 \\
            \bottomrule
        \end{tabular}
    }
    \caption{Summary of Post-hoc Dunn's Test results for sentence position impact on ROUGE 1, BERTScore, and BLEURT metrics. No significant differences observed.}
    \label{tab:res_sens_sent_id}
\end{table}

The analysis was split into two main tests: the Kruskal-Wallis H-Test to identify overall differences across sentence positions and the Post-hoc Dunn's Test to investigate pairwise differences between sentence positions.

The Kruskal-Wallis H-Test was applied to compare the distributions of scores for ROUGE 1, BERTScore, and BLEURT across three sentence positions (beginning, middle, end) within clinical notes from the validation sets of MS and UW.
As shown in Table~\ref{tab:res_sens_sent_id}, statistically significant differences were found in the MS dataset for ROUGE 1 and BLEURT metrics, suggesting sensitivity to sentence positioning.

Following the Kruskal-Wallis H-Test, a Post-hoc Dunn's Test was performed to conduct pairwise comparisons between sentence positions for each evaluation metric.
The Post-hoc Dunn's Test revealed no statistically significant differences between any pairwise comparisons of sentence positions for all evaluated metrics, suggesting that while overall differences exist, specific pairwise comparisons did not reach statistical significance.

\end{document}